\pdfoutput=1

\documentclass[11pt]{article}

\usepackage{acl}
\usepackage{titlesec}
\usepackage{times}
\usepackage{pgfplots}
\usepackage{latexsym}
\usepackage{booktabs}
\usepackage{multirow}
\pgfplotsset{compat=1.18}
\usepackage{amssymb}
\usepackage{pifont}
\usepackage{longtable}
\usepackage{array}
\usepackage{adjustbox}
\usepackage[T1]{fontenc}
\usepackage[utf8]{inputenc}
\usepackage{microtype}
\usepackage{inconsolata}
\usepackage{graphicx}

\usepackage[utf8]{inputenc}
\usepackage{textgreek}

\usepackage{fvextra}            
\usepackage{listings}           
\usepackage[most]{tcolorbox}
\tcbuselibrary{skins,breakable,listings}
\usepackage{tikz}
\usetikzlibrary{shapes.geometric} 

\tikzset{
    startstop/.style = {ellipse, minimum width=2.5cm, minimum height=1cm,text centered, draw=black, fill=gray!20},
    process/.style = {rectangle, rounded corners, minimum width=4cm, minimum height=1cm,text centered, draw=black, fill=blue!20},
    api/.style = {rectangle, rounded corners, minimum width=4cm, minimum height=1cm,text centered, draw=black, fill=yellow!20},
    decision/.style = {diamond, minimum width=3cm, minimum height=1cm,text centered, draw=black, fill=green!20},
    arrow/.style = {thick,->,>=stealth}
}

\newtcolorbox{PromptBox}[2][]{%
  enhanced,
  breakable,
  colback=gray!3,
  colframe=black!20,
  title={#2},
  fonttitle=\bfseries,
  listing only,
  listing engine=listings,
  listing options={
    basicstyle=\ttfamily\small,
    breaklines=true,
    breakatwhitespace=false,
    columns=fullflexible,
    upquote=true,
    showstringspaces=false,
    tabsize=2
  },
  #1
}

\title{Deflanderization for Game Dialogue: Balancing Character Authenticity with Task Execution in LLM-based NPCs}

\author{
\textbf{Pasin Buakhaw\textsuperscript{1}\thanks{Equal contribution.}},
\textbf{Kun Kerdthaisong\textsuperscript{2}\footnotemark[1]},
\textbf{Phuree Phenhiran\textsuperscript{2}\footnotemark[1]},\textbf{Pitikorn Khlaisamniang\textsuperscript{3}},\\
\textbf{Supasate Vorathammathorn\textsuperscript{3}},\textbf{Piyalitt Ittichaiwong\textsuperscript{4,5}\thanks{Corresponding authors.}},
\textbf{Nutchanon Yongsatianchot\textsuperscript{2}\footnotemark[2]}\\[4pt]
\textsuperscript{1}Department of Computer Engineering and Digital Technology, Faculty of Engineering, Chulalongkorn University\\
\textsuperscript{2}Faculty of Engineering, Thammasat School of Engineering, Thammasat University\\
\textsuperscript{3}Artificial Intelligence Association of Thailand\\
\textsuperscript{4}School of Biomedical Engineering \& Imaging Sciences, King’s College London\\
\textsuperscript{5}Siriraj Informatics and Data Innovation Center (SIData+), Faculty of Medicine, Siriraj Hospital, Mahidol University
}

\begin{document}
\maketitle
\vspace{-2em} 

\begin{abstract}
The emergence of large language models (LLMs) has opened new opportunities for creating dynamic non-player characters (NPCs) in gaming environments, enabling both functional task execution and persona-consistent dialogue generation. In this paper, we (\textbf{TU\_Character\_lab}) report our participation in the Commonsense Persona-Grounded Dialogue Challenge (CPDC) 2025 Round 2, which evaluates agents across three tracks: task-oriented dialogue, context-aware dialogue, and their integration. Our approach combines two complementary strategies: (\textit{i}) lightweight prompting techniques in the API track, including a \textbf{Deflanderization} prompting method to suppress excessive role-play and improve task fidelity, and (\textit{ii}) fine-tuned large models in the GPU track, leveraging Qwen3-14B with supervised finetuning (SFT) and Low-Rank Adaptation (LoRA). Our best submissions ranked \textbf{2\textsuperscript{nd}} on Task~1, \textbf{2\textsuperscript{nd}} on Task~3 (API track), and \textbf{4\textsuperscript{th}} on Task~3 (GPU track).

\end{abstract}

\section{Introduction}
\label{Introduction}

The revolution of large language models (LLMs) has demonstrated that transformer architectures can engage in human-like dialogue interactions within virtual environments. Recent studies have categorized persona-enabled LLMs into two distinct adaptation approaches: user-focused personalization and environment-based role-playing~\citep{tseng2024two}.

First, user persona-LLMs are designed as purpose-built assistants that adapt to individual users' preferences, backgrounds, and behavioral patterns ~\citep{salemi2024lamp}. These personalization systems leverage user-specific information to provide tailored responses, recommendations, and interactions. For example, LaMP (Large Language Models Meet Personalization) introduces comprehensive benchmarks for evaluating personalized text generation ~\citep{salemi2024lamp}, while another work explores personalized dialogue agents that maintain consistent user preferences across conversations ~\citep{zhang2023personachat}.

Second, environment adaptation involves LLMs tasked with maintaining consistent personas within specific contexts, commonly referred to as role-playing. This approach has gained significant traction in multi-agent systems where LLMs assume distinct professional roles. ChatDev ~\citep{qian2023chatdev} exemplifies this paradigm by creating a virtual software development company where different agents handle specialized tasks such as programming, testing, and documentation. Similarly, MetaGPT ~\citep{hong2023metagpt} proposes a meta-programming framework for collaborative multi-agent workflows, while Generative Agents ~\citep{park2023generative} demonstrates believable human behavior simulation through persistent agent personas. Advanced frameworks like CAMEL ~\citep{li2023camel} and Voyager ~\citep{wang2023voyager} further explore how role-playing agents can engage in complex problem-solving and open-ended exploration tasks.

These developments showcase the remarkable ability of modern LLMs to facilitate and embody given personas, with applications spanning from personalized user assistance to sophisticated multi-agent collaborations in virtual environments ~\citep{jiang2023evaluating}.

\begin{figure}
    \centering
    \includegraphics[width=1.02\linewidth]{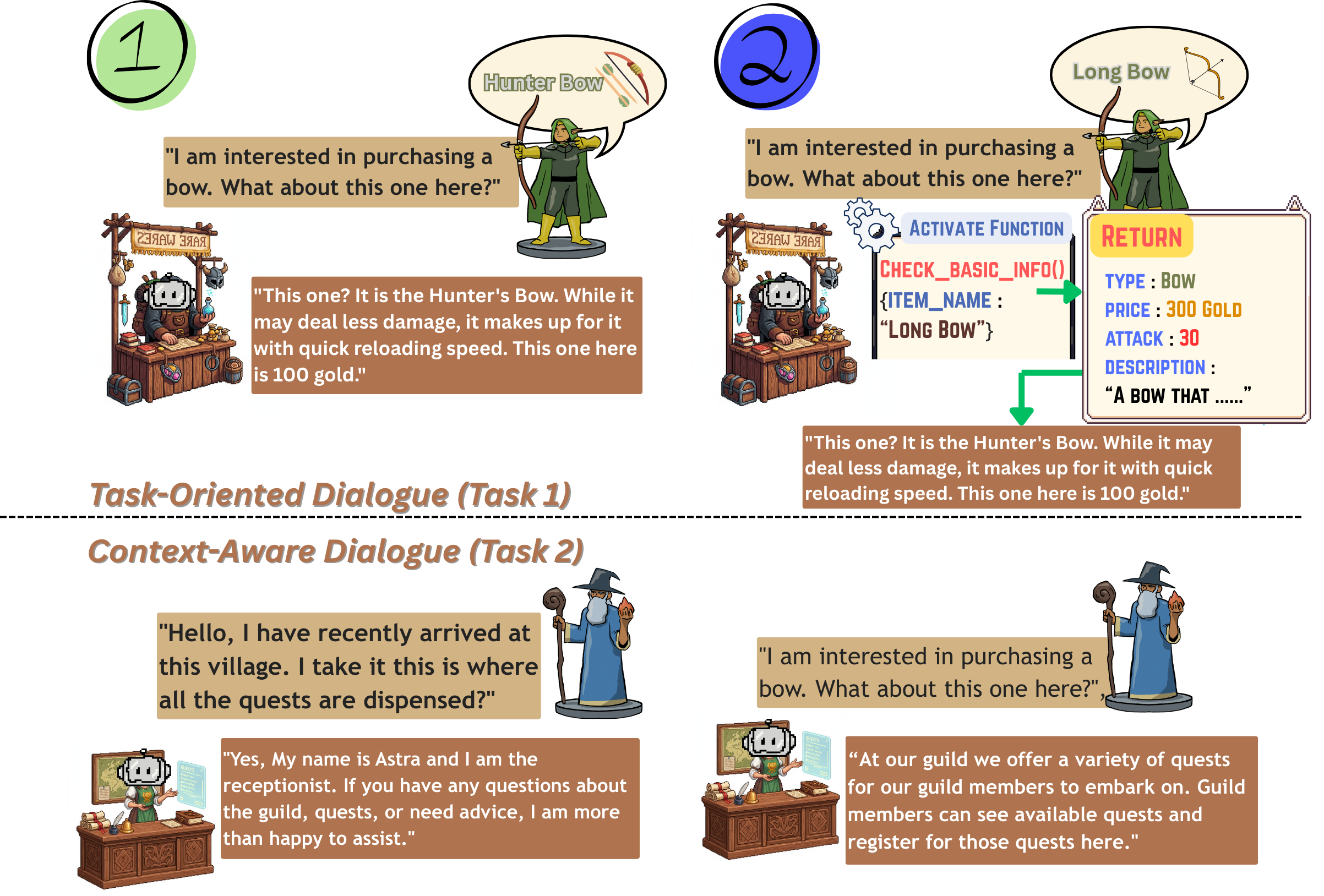}
    \caption{Examples of player–NPC interactions using LLM-based agents in the CPDC 2025 competition, 
    Top panel: Early Summer 7 PM, clear night at the Weapon Shop, showing an example of user-NPC interaction in \textbf{Task 1} (function generation). Bottom panel: Late Winter 2 PM, rainy conditions at the Quest Reception Desk, showing an example of dialogue generation in \textbf{Task 2}.}
    \label{fig:example}
\end{figure}

Despite the rapid growth of LLM research, the entertainment field has remained relatively underexplored, particularly in traditional entertainment media creation such as video games. Conventional game development relies heavily on programmed logic, where in-game events and character interactions follow predetermined scripts and dialogue trees. To enhance player immersion and narrative depth, developers have begun incorporating LLMs as integral components of NPCs. This integration enables them to exhibit human-like behaviors and engage in dynamic, contextually-aware conversations with players ~\citep{song2024llm}.

However, maintaining the consistency and depth of these dynamic personas over long-term interactions presents a significant challenge. One such pitfall, drawn from media analysis, is the trend of "flanderization"~\citep{Larsen2019Gatekeeping}. Flanderization is the process through which a complex character is progressively simplified over time, eventually becoming a caricature defined by a single, exaggerated trait. The term originates from the character Ned Flanders in The Simpsons, who evolved from a genuinely good-natured neighbor—whose faith was one of many aspects of his personality—into a one-dimensional religious zealot.

Recent advances in LLM-driven NPCs demonstrate significant potential for transforming player experiences. Cross-platform dialogue systems allow NPCs to maintain consistent interactions across both game environments and social platforms like Discord ~\citep{song2024llm}, creating unprecedented continuity in player-character relationships. Collaborative quest completion systems in Minecraft showcase how LLM-driven NPCs can work alongside human players to accomplish shared objectives ~\citep{rao2024collaborative}, while function-calling capabilities enable AI Game Masters or NPCs to manage complex game mechanics and narrative progression \cite{song2024thirteenhourssolvelabyrinth}. Furthermore, specialized datasets like MCPDial ~\cite{alavi2024mcpdialminecraftpersonadrivendialogue} and PeaCoK ~\citep{gao-etal-2023-peacok} provide rich persona-driven dialogue collections that enhance the authenticity and depth of NPC interactions, supporting the development of more sophisticated conversational agents in gaming environments.

The growing interest in persona-grounded gaming applications has culminated in organized initiatives such as the Commonsense Persona-Grounded Dialogue Challenge (CPDC) 2025~\citep{cpdc2025}. This competition invited submissions aimed at developing NPC agents capable of demonstrating both persona consistency and task execution proficiency within a fantasy Role-Playing Game (RPG) environment, as illustrated in Figure~\ref{fig:example}.

Building upon this challenge, our work investigates multiple complementary strategies for enhancing the coherence and reliability of LLM-driven NPCs across diverse interaction settings. Specifically, we explore a \textit{Deflanderization prompting} approach that mitigates character drift and preserves personality coherence during extended dialogues, ensuring balanced performance between dialogue generation and function execution. To further strengthen contextual grounding, we integrate a \textit{Retrieval-Augmented Generation (RAG)} mechanism that incorporates memory and similarity-based retrieval from prior interactions, allowing NPCs to produce responses aligned with both in-game history and established world knowledge. Finally, we employ \textit{Supervised Finetuning (SFT)} with Low-Rank Adaptation (LoRA) to refine model behavior at the parameter level, enhancing stylistic consistency and functional precision while remaining computationally efficient.

Together, these methods constitute a unified framework that examines the interplay between prompting, retrieval-augmented reasoning, and finetuned adaptation in achieving persona-consistent, context-aware, and goal-directed NPC performance within the CPDC 2025 setting.

From our participation in this challenge across every track, both GPU and API divisions, the following are key points that we investigated:
\begin{itemize}
    \item \textbf{Deflanderization prompting technique to maintain dialogue generation and function generation ability in common fantasy RPG world setting.}
    \item \textbf{Explore the performance trade-offs between dialogue generation and function generation tasks using the proposed prompt engineering technique.}
    
\end{itemize}


\section{Related Work}
\subsection{Agents for Game-Oriented Dialogue}
\label{Agents for Oriented-Game-Dialogue}

Task-oriented systems are designed to efficiently complete specific tasks within larger workflows, often serving as prerequisites for later stages. Integrating agentic systems enhances these workflows by enabling agents to analyze problems, plan, and execute actions toward defined goals. Research on task-oriented dialogue (TOD) systems, such as ~\citep{kazi_useragent_tod}, benchmarks agent performance by assessing planning effectiveness, goal alignment, and interaction methods used to gather information and achieve successful outcomes.

In the context of gaming, completing a sequence of events often involves accomplishing a series of tasks. To aid players, especially newcomers, ~\citep{lee-etal-2025-aman} developed a specialized game assistant. This assistant leverages an LLM that has undergone continuous pre-training and instruction tuning to answer specific game-related questions, thereby helping users navigate complex game mechanics.

To ensure that interactive agents can successfully complete their objectives within a game~\citep{adon_goalgame} introduced a framework that utilizes two distinct agents: a Dialogue agent and a goal-verifying agent. This system employs shared memory to manage interactions, ensuring that dialogue and actions remain aligned with the overarching task goals.

\subsection{Tool calling}
\label{Tool calling agent}

Tool-calling or function-calling, an ability of LLMs to interact with external tools or functions, experienced a recent surge in interest, driven by the potential of LLMs to autonomously complete tasks by dynamically accessing and acting upon external resources, extending their capabilities to become agentic AI ~\citep{Xu2025,patil2025bfcl}.

The architecture of these agents typically involves a multi-step framework to ensure accuracy in complex, real-world tasks. This framework includes components for executing actions, perceiving the environment, validating results, controlling the overall plan, and retrieving tools from a toolset ~\citep{Xu2025}. 

A key challenge in this domain is the development of robust evaluation benchmarks. While existing benchmarks have focused on single-control environments where only the AI agent can interact with tools, recent work has introduced more complex scenarios. For instance ~\citep{barres_tau2}, the $\tau^2$-Bench introduces a dual-control environment where both the agent and the user can utilize tools to act in a shared, dynamic world. This setup is designed to more accurately represent real-world collaborative scenarios, such as technical support, and to expose the challenges of agent coordination and communication that are absent in single-user control evaluations. The performance of LLMs degrades significantly in such dual-control settings, underscoring the difficulty of guiding user actions and the importance of further research in this area.

\section{Competition Overview}

\subsection{Competition Tasks}

The CPDC competition aims to facilitate dialogues that seamlessly integrate contextual understanding, knowledge utilization and task execution capabilities in a fantasy RPG game setting \citep{cpdc2025}. The competition comprises two tracks, API Track and GPU Track (detailed in the next section), and each track consists of three tasks:
\begin{itemize}
  \item \textit{Task 1: Task-Oriented Dialogue Agents}, 
  \item \textit{Task 2: Context-Aware Dialogue Agents}, 
  \item \textit{Task 3: Integrated Contextual Dialogue and Task Execution (combining both Task 1 and Task 2)}. 
\end{itemize}

Examples of these tasks are illustrated in Figure~\ref{fig:example}.

\subsubsection{Task 1: Task-Oriented Dialogue Agents}
\label{Task 1: Task-Oriented Dialogue Agents}

In this task, participants develop dialogue response generation systems that operate in two phases: first, assessing conversational context to determine necessary function calls, and second, executing these calls with appropriately selected arguments that align with the conversation for task execution. For example, merchant NPCs in games select weapons to sell based on player dialogue. Evaluation in this track primarily focuses on the correctness of function calls and the accuracy of argument selection.

\subsubsection{Task 2: Context-Aware Dialogue Agents}
\label{Task 2: Context-Aware Dialogue Agents}

In this task, participants develop dialogue response generation systems that focus on generating NPC responses with tones aligned to their assigned personas. Evaluation emphasizes the extent to which generated responses maintain consistency with the NPC's defined persona and character traits.

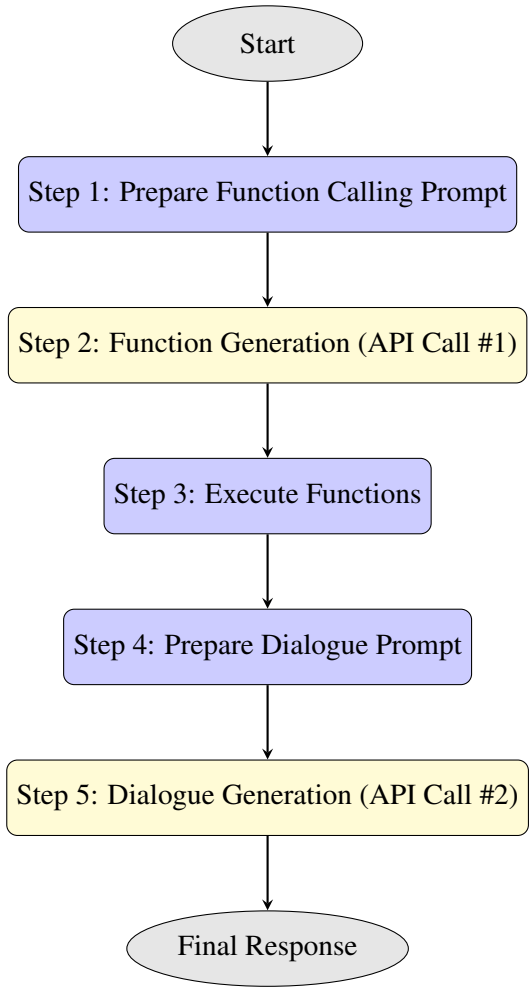
\begin{figure}[ht]
    \centering
    \begin{tikzpicture}[node distance=2cm]
    \node (start) [startstop] {Start};
    \node (step1) [process, below of=start] {Step 1: Prepare Function Calling Prompt};
    \node (step2) [api, below of=step1] {Step 2: Function Generation (API Call \#1)};
    \node (step3) [process, below of=step2] {Step 3: Execute Functions};
    \node (step4) [process, below of=step3] {Step 4: Prepare Dialogue Prompt};
    \node (step5) [api, below of=step4] {Step 5: Dialogue Generation (API Call \#2)};
    \node (end) [startstop, below of=step5] {Final Response};
    
    \draw [arrow] (start) -- (step1);
    \draw [arrow] (step1) -- (step2);
    \draw [arrow] (step2) -- (step3);
    \draw [arrow] (step3) -- (step4);
    \draw [arrow] (step4) -- (step5);
    \draw [arrow] (step5) -- (end);

    \end{tikzpicture}
    \caption{Main pipeline for the API Track task 3. The prompting stages are \textbf{Step1} and \textbf{Step4} and generataion stages are in \textbf{Step2} and \textbf{Step5}}
\label{fig:api_pipeline}
\end{figure}

\section{Competition Tracks}
\label{Competition Tracks}

\subsection{API Track}
\label{API Track}

In the API track, participants submit their work within specific environment and constraints such as the allowed LLM is GPT-4o-mini (see \ref{appendix:API Track} for full details). We focused on prompting methods. Our pipeline is illustrated in Figure \ref{fig:api_pipeline}. We systematically explored the following prompting approaches:

\begin{itemize}
    \item \textbf{D (Deflanderization):}\label{key:D} Prompts the model to respond naturally and concisely while avoiding exaggerated role-playing. Our error analysis of the baseline setup from the challenge comparing generated responses with gold-standard outputs revealed that the baseline setup often produced overly elaborate and contextually diffuse outputs, focusing excessively on the narrative setting (e.g., adopting an RPG character persona) rather than addressing the immediate conversational intent to player. 
In contrast, the gold responses reflected a more human-like understanding of player requests and directly activated the appropriate functions with clarity.

    \item \textbf{F (Fewshot):}\label{key:F} Includes two sample dialogues (merchant and guild receptionist) from \texttt{sample.json} in the prompt.

    \item \textbf{ZeroShot:}\label{key:Zero} Uses the initial baseline prompt from the competition repository.
    
    \item \textbf{CoT (Chain of Thought):}\label{key:CoT} Instructs the model to think step-by-step before answering.

    \item \textbf{RW (Remove world setting):}\label{key:RW} Removes worldview information when constructing dialogue prompts.

    \item \textbf{G (Guide):}\label{key:G} Guides response style by limiting to 1–2 short sentences, using simple language, and restricting to provided knowledge.

    \item \textbf{MW (Most word):}\label{key:MW} Guides word usage and provides example phrases.

    \item \textbf{Define function:}\label{key:Func} Provides two sample function arguments (merchant and guild receptionist) with their items in JSON format.
\end{itemize}

Our best submission (\textbf{ranked 2\textsuperscript{nd} on Task 3, 2\textsuperscript{nd} on Task 1 and 5\textsuperscript{th} on Task 2}) on public leader board used only \textbf{D-RW} combined with two turns of sample dialogues. 

\subsection{GPU Track}
\label{GPU Track}

Due to the compute limitations described in Appendix~\ref{appendix:Compute Constraints}, we selected models that can be executed on the \textit{AWS g5e.2xlarge instances with L40s GPUs} instance. We first validated inference submission feasibility using Qwen2.5~\citep{Qwen_AnYang}, Qwen3~\citep{Qwen3_an_yang}, LLaMA3.1~\citep{llama31_aron}, and Phi-4~\citep{phi4_marah}, before proceeding with finetuning experiments on both initial and augmented data.

To improve dialogue grounding, we incorporated a hybrid \textbf{Retrieval Augmented Generation (RAG) + Memory} approach. The retrieval module encodes both player and NPC conversation histories using Qwen3-Embedding-0.6B, enabling similarity search across pre-collected interaction datasets. The retrieved context is injected at two stages: (i) \textit{Function Selection}, where prior conversations guide accurate tool invocation, and (ii) \textit{Dialogue Drafting}, where relevant NPC responses provide style and factual grounding. 

Additionally, we explored a \textbf{RAG+Refine} step, where generated drafts are rewritten to match the tone and length of high-similarity golden responses, ensuring stylistic consistency with provided in-game dialogue.

Our best-performing submission (\textbf{ranked 4\textsuperscript{th} on Task 3 public leaderboard}) was achieved  with Qwen3-14B. We applied Supervised Finetuning (SFT) with Low-Rank Adaptation (LoRA)~\citep{hu2022lora} using the Unsloth framework~\citep{unsloth}. The training procedure was divided into two stages: \textbf{(1) Full SFT} on initial and synthetic multi-turn dialogue data, followed by \textbf{(2) LoRA-SFT} (rank=32, $\alpha=32$) on combined dialogue and function-calling datasets. 

We generated the datasets using \textbf{gemini-2.5-pro-preview-05-06}~\citep{google2025gemini2.5} for function-calling data and \textbf{GPT-4o-mini}~\citep{openai_gpt4o_mini} for dialogue data. The generated datasets consist of: Multi-turn (2,800 data points), Multi-turn reasoning (2,800 data points) for \textbf{Task 2} (\ref{Task 2: Context-Aware Dialogue Agents}) and Funtion-calling generation (328 data points) for \textbf{Task 1}~(\ref{Task 1: Task-Oriented Dialogue Agents}). Prompts used for data generation are provided in ~\ref{appendix:Additional Data Generation}.

For inference, we optimized deployment with vLLM~\citep{kwon2023vllm} using the following hyperparameters:
dtype='bfloat16', gpu\textunderscore memory\textunderscore utilization=0.8,
enable\textunderscore LoRA, max\textunderscore model\textunderscore len=4096,
and disable\textunderscore sliding\textunderscore window=True,
enabling Qwen3-14B to run within the L40s memory budget.



\section{Results}
\label{Results}

\begin{table*}[ht]
    \centering
    \small
    \setlength{\tabcolsep}{1mm}
    \caption{\textbf{API Track} Task 1 Result}
    \begin{tabular}{@{}ccccccccc@{}}
        \toprule
        Dataset & metrics & ZeroShot$^{\hyperref[key:Zero]{\text{(Z)}}}$  & CoT$^{\hyperref[key:CoT]{\text{(CoT)}}}$ & F$^{\hyperref[key:F]{\text{(F)}}}$ & Define function$^{\hyperref[key:Func]{\text{(func)}}}$ &\textbf{Our Best$^{\hyperref[key:D]{\text{(D)}},\,\hyperref[key:RW]{\text{(RW)}}}$}\\ 
        \specialrule{1.5pt}{0pt}{0pt}
        
        \multirow{3}{*}{\texttt{train.json}}
        & Function name exact match & 0.622 & 0.537 & \underline{0.633} & 0.615 &\textbf{0.714}\\ 
        & Function argument exact match  & \underline{0.226} & 0.211 & 0.199 & 0.210 &\textbf{0.359}\\ 
        & BERTScore  & 0.542 & \underline{0.566} & 0.538 & 0.539& \textbf{0.569} \\ 
        \midrule

        \multirow{3}{*}{\texttt{sample.json}}
        & Function name exact match & 0.667 & 0.333 & 0.600 & \underline{0.714} & \textbf{0.727}\\ 
        & Function argument exact match  & 0.333 & 0.000 & 0.100 & \textbf{0.429} &\underline{0.364}\\ 
        & BERTScore  & \underline{0.509} & \textbf{0.534} & 0.491 & 0.496 &\textbf{0.534}\\ 
        \midrule

        \multirow{1}{*}{test(submission)} 
        
        & CPDCscore(Task 1)  & 0.422 & 0.383 & \underline{0.441} & 0.430 & \textbf{0.586}\\ 
        \bottomrule
        
    \end{tabular}
    \label{tab: api track task1}
\end{table*}

\begin{table*}[ht]    
    \centering
    \small
    \setlength{\tabcolsep}{1mm}
    \caption{\textbf{API Track} Task 3 Result}
    \begin{tabular}{@{}cccccccc@{}}
        \toprule
        Dataset & metrics 
        & ZeroShot
        & D$^{\hyperref[key:D]{\text{(D)}}}$ 
        & D-F-RW$^{\hyperref[key:RW]{\text{(RW)}}}$ 
        & D-F-G-RW$^{\hyperref[key:G]{\text{(G)}}}$ 
        & D-F-MW-G-RW$^{\hyperref[key:MW]{\text{(MW)}}}$ 
        & \textbf{Our Best} \\ 
        \specialrule{1.5pt}{0pt}{0pt}
        
        \multirow{3}{*}{\texttt{train.json}}
        &BLEU-4& 0.031 & 0.035 & 0.032 & \underline{0.041} & \textbf{0.043} &0.040\\ 
        &Word-level F1& \underline{0.293} & 0.273 & 0.279 & 0.290 & \textbf{0.300}&\underline{0.292}\\ 
        &BERTScore& 0.525 & \textbf{0.543} & 0.527 & 0.537 & \underline{0.542}&0.540\\ 
    
        \midrule
        
        \multirow{3}{*}{\texttt{sample.json}} 
        &BLEU-4& 0.027 & 0.028 & 0.024 & 0.030 & \underline{0.050}&\textbf{0.095}\\ 
        &Word-level F1& 0.276 & 0.281 & 0.293 & \underline{0.319} & \underline{0.327}&\textbf{0.331}\\ 
        &BERTScore& 0.536 & 0.548 & 0.553 & \underline{0.557} & \textbf{0.568}&\underline{0.577}\\ 
        \bottomrule

        \multirow{3}{*}{test(submission)} 
        &CPDCscore(Task 1)& 0.422 & 0.436 & 0.429 & 0.432 & \underline{0.441}& \textbf{0.587}\\
        &CPDCscore(Task 2)& 0.598 & 0.614 & \textbf{0.619} & 0.612 & 0.612&\underline{0.615}\\
        &CPDCscore(all)& 0.510 &0.525 & 0.524 & 0.522 & \underline{0.526}& \textbf{0.601}\\
        
        \bottomrule
    \end{tabular}
    \label{tab: api track task3}
\end{table*}

\subsection{API Track}

Before submitting to the ~\citep{aicrowd} submission system, we focused on testing the \textbf{API track} on existing datasets to explore possible prompting technique. The dataset consists of \textbf{Task 1} \texttt{train.json}, \texttt{sample.json} and \textbf{Task 2}: \texttt{train.json}, \texttt{sample.json}. Tables~\ref{tab: api track task1} and~\ref{tab: api track task3} summarize the  the API track results for Task 1 and Task 3, respectively.  
We observe several notable trends:

\textbf{1. Effectiveness of Deflanderization  prompting.}  

Across both tasks, the Deflanderization  (D) strategy consistently improved scores compared to the zero-shot baseline. In Task~3 (Table~\ref{tab: api track task3}), D-RW achieved a +0.013 absolute gain in CPDCscore (all) compared to zero-shot. This supports our hypothesis that overly strong role-playing can hinder functional correctness by diverting the model toward stylistic embellishment rather than more realistic character.

\textbf{2. Sample-based prompting further boosts accuracy.}  


Adding few-shot examples (F) to the Deflanderization prompt provided clear benefits in Task 1 (Table~\ref{tab: api track task1}), with improvements of +0.092 and +0.133 on \texttt{train.json}, respectively. Notably, our best-performing API submission combined \textbf{D-RW with two-turn few-shot examples}, yielding the highest leaderboard placement (\textbf{2\textsuperscript{nd}} on Task 3, \textbf{2\textsuperscript{nd}} on Task 1, and \textbf{5\textsuperscript{th}} on Task~2).  

\textbf{3. Limited benefits of more complex prompting.}  

Chain-of-Thought (CoT), guiding responses (G), and Most Word (MW) constraints yielded marginal or inconsistent gains. For instance, CoT improved BERTScore in Task 1 but decreased function argument accuracy, likely due to verbose reasoning diluting key arguments. Similarly, MW improved BLEU on \texttt{train.json} but did not transfer to the leaderboard CPDCscore. This suggests that lightweight strategies (D + few-shot) are more robust under competition constraints than complex, multi-signal prompts for these tasks.

Table~\ref{GPU Track on task3} presents results for Task~3 under the GPU track.  

\begin{table*}[ht]
    \centering
    \caption{Result submission at \textbf{GPU Track} on Task 3.}
    \label{GPU Track on task3}
    \setlength{\tabcolsep}{1.4mm}
    \begin{tabular}{@{}|c|c|c|c|c|@{}}
        \toprule
        Model  & Method & Score Task 1 & Score Task 2 & All\\ 
        \specialrule{1.5pt}{1.5pt}{1.5pt}
        \multirow{1}{*}{LLaMA3.1-8B} 
          & baseline & 0.439& 0.333& 0.386\\
        \midrule
        \multirow{1}{*}{Phi4-mini} 
          & baseline & 0.328& 0.354& 0.341\\
        \midrule
        \multirow{1}{*}{Qwen2.5-7B} 
          & baseline & 0.440& \underline{0.587}& 0.513\\
        \midrule
        \multirow{1}{*}{Qwen3-8B} 
          &baseline& 0.449& \underline{0.587}& 0.518\\
        \midrule
          \multirow{3}{*}{Qwen3-14B-FP8}
          &Rag + Refine & \underline{0.522}& 0.549& \underline{0.535}\\
          &Rag Memory& 0.502& 0.532& 0.517\\ %
          &SFT + LoRA~\textbf{(Our Best)}& \textbf{0.590}& \textbf{0.606} & \textbf{0.598}\\
        \bottomrule
    \end{tabular}
\end{table*}

\textbf{1. Model scaling and finetuning are critical.}  Baseline submissions with smaller models (e.g., LLaMA3.1-8B, Phi-4-mini) underperformed, with all-scores below 0.40. In contrast, Qwen3-14B with full SFT and LoRA achieved a significant improvement, reaching \textbf{0.598 all-score}, ranking \textbf{4\textsuperscript{th}} on the leaderboard. This highlights the importance of both model size and targeted finetuning on domain-specific data.  

\textbf{2. Retrieval augmentation provided modest improvements.}  RAG+Refine and RAG+Memory approaches improved Qwen3-8B performance to 0.522 for Task 1, showing that retrieval helps stabilize dialogue grounding. However, these methods fell short of the gains achieved by LoRA-SFT. We attribute this to the limited scale of the retrieval corpus and the challenge of injecting retrieved context seamlessly without overloading prompts.  

\textbf{3. Trade-off between Task 1 and Task 2.}  Interestingly, while RAG+Refine gave the best Task~1 score (0.522), it underperformed on Task 2 compared to baseline. Conversely, LoRA-SFT balanced both tasks, producing the highest joint score. This suggests that alignment between functional reasoning (Task 1) and persona-grounded dialogue (Task 2) requires joint optimization, rather than modular improvements in isolation.  

\section{Discussion}

Overall, our findings reveal complementary strategies across the API and GPU tracks. Prompting-based Deflanderization with few-shot grounding proved effective in low-resource API settings, while finetuned large models dominated the GPU track. Importantly, both tracks highlighted the challenge of balancing persona consistency with functional precision: methods that improved role-play fidelity sometimes hurt argument correctness, and vice versa. Future work should explore hybrid strategies that unify lightweight prompting with retrieval-augmented finetuning, enabling agents to sustain both accuracy and believability in fantasy RPG environments. Our final rankings are in Appendix~\ref{Final Leader Board}.

\section*{Acknowledgments}
This research was supported by the Faculty of Engineering, Thammasat School of Engineering, Thammasat University also thanks to PreceptorAI that provides API for generate additional training data.

\bibliography{custom}

\clearpage
\appendix
\section*{Appendix}
\addcontentsline{toc}{section}{Appendix} 

\section{Exploratory Data Analysis}
\label{Exploratory Data Analysis}
Before doing some experiments, we perform data analysis on \textbf{Task 1\_train}.\textbf{json} and \textbf{Task 2\_train}.\textbf{json}.

\begin{figure}[ht]
    \centering
    \includegraphics[width=1.0\linewidth]{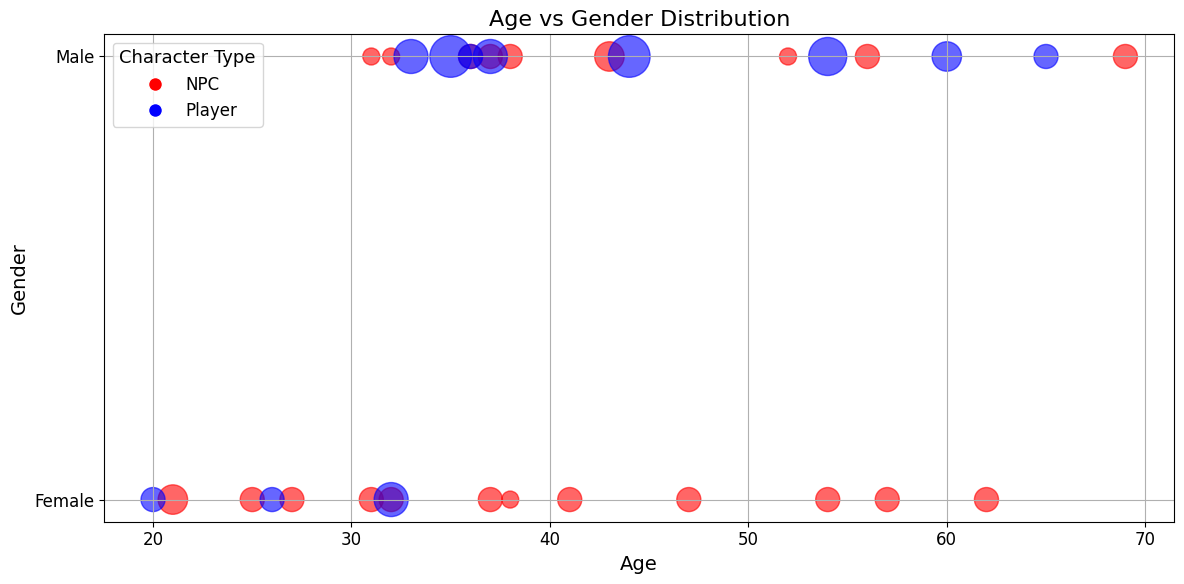}
    \caption{Age-gender of characters in Task 2\_\texttt{train.json}, the diagram shown that balanced NPC characters(20 merchant and 20 guild receptionist) most NPC are women with the younger age than men.}
    \label{appendix:age-gender}
\end{figure}

\begin{figure}[ht]
    \centering
    \includegraphics[width=1.0\linewidth]{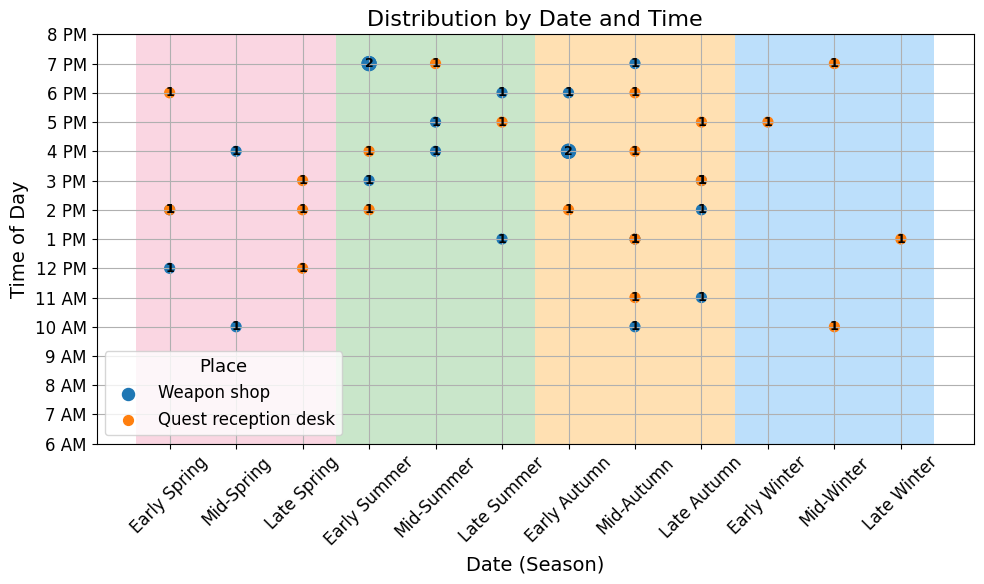}
    \caption{Date-time distribution in Task 2\_\texttt{train.json}, most of event occur after 1 pm and there are only quest reception place event in winter season.}
    \label{appendix:Date-time}
\end{figure}

\begin{figure}[ht]
    \centering
    \includegraphics[width=1.0\linewidth]{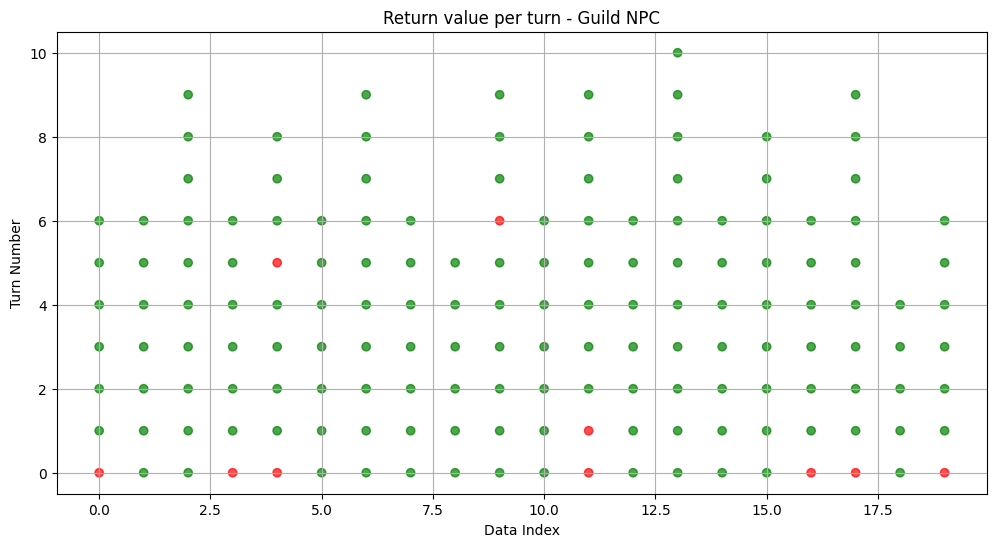}
    \caption{Guild NPC Response Return Value Ratios in Task 1\_\texttt{train.json}
(Green = return; Red = no return)}
    \label{fig:enter-label}
\end{figure}

\begin{figure}[ht]
    \centering
    \includegraphics[width=1.0\linewidth]{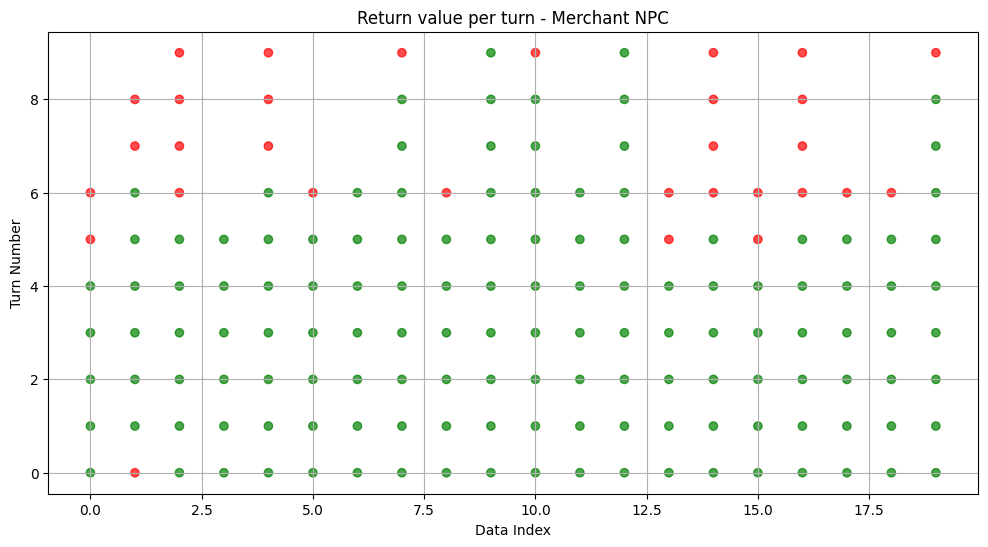}
    \caption{Merchant NPC Response Return Value Ratios in Task 1\_\texttt{train.json}
(Green = return; Red = no return)}
    \label{fig:enter-label2}
\end{figure}

\begin{figure}[ht]
    \centering
    \includegraphics[width=1.0\linewidth]{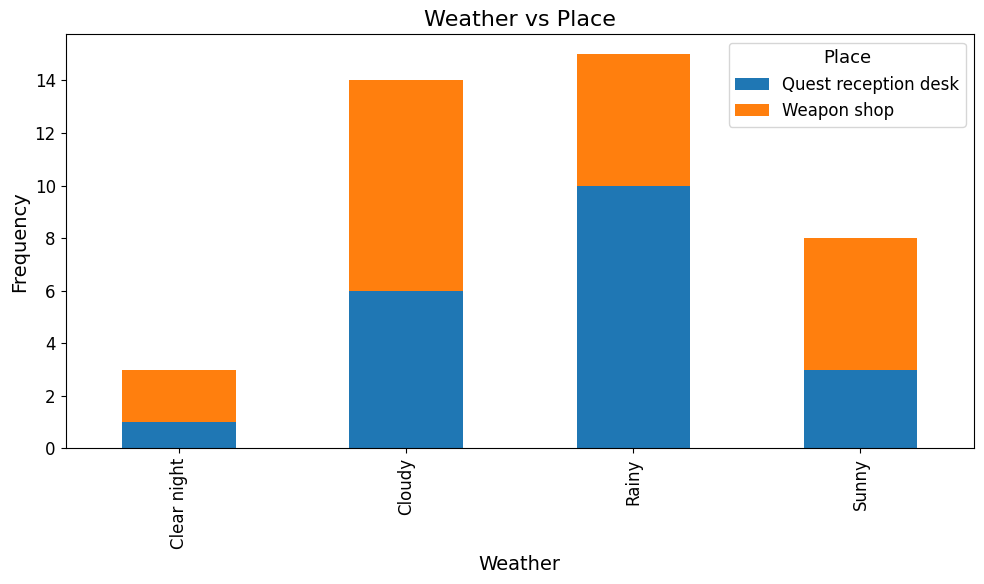}
    \caption{Barplot of frequency merchant/guild receptionist mapped with their weather on that situation.}
    \label{fig:enter-label3}
\end{figure}

\section{Evaluation Metrics}
\label{appendix:Evaluaiton Metrics}
In CPDC2023~\citep{cpdc2023} they used WordF1, BLEU, CPDScore, USEScore and BERTScore to automatically evaluate the dialogue generation so we try to use some of these metrics in our local environment for task dialogue generation~\ref{appendix:task 2 Metrics} and task function generation we use these metrics in experiments \ref{appendix:task 1 Metrics}.

While automatic metrics alone are not fully reliable for evaluating dialogue systems~\citep{liu-etal-2016-evaluate, novikova-etal-2017-need}, the organizers therefore relied on human evaluation for the final private leaderboard.

\subsection{Task 1}
\label{appendix:task 1 Metrics}
\subsubsection{Function name exact match}
This metric checks if the predicted function name matches the reference exactly:
\begin{equation}
\text{Acc}_{name} = \frac{1}{N} \sum_{i=1}^{N} \mathbf{1}\{f^{pred}_i = f^{ref}_i\},
\end{equation}
where $f^{pred}_i$ and $f^{ref}_i$ denote the function's name sets of instance i.

\subsubsection{Function argument exact match}
This metric checks if all predicted arguments exactly match the reference:
\begin{equation}
\text{Acc}_{args} = \frac{1}{N} \sum_{i=1}^{N} \mathbf{1}\{A^{pred}_i = A^{ref}_i\},
\end{equation}
where $A^{pred}_i$ and $A^{ref}_i$ denote the argument sets of instance $i$.

\subsubsection{BERTScore}
\label{appendix:BERTScore}
We also measure semantic similarity of function calls with BERTScore~\citep{BERTscore}.  
Given tokens $X=(x_1,\dots,x_m)$ from prediction and $Y=(y_1,\dots,y_n)$ from reference:
\begin{equation}
s(x_i,y_j) = \frac{E(x_i)\cdot E(y_j)}{\|E(x_i)\| \|E(y_j)\|},
\end{equation}
\begin{equation}
P = \tfrac{1}{m} \sum_{i=1}^{m} \max_{j} s(x_i,y_j),
\\
R = \tfrac{1}{n} \sum_{j=1}^{n} \max_{i} s(y_j,x_i)
\end{equation}
\begin{equation}
\text{BERTScore-F1} = \frac{2PR}{P+R}.
\end{equation}

\subsection{Task 2}
\label{appendix:task 2 Metrics}
In Track 2, we evaluate role-playing consistency using four metrics, including BERTScore (described in Appendix~\ref{appendix:BERTScore}), with the remaining metrics detailed below.

\subsubsection{BLEU-4}
BLEU-4 is based on modified $n$-gram precision (for $n=1,2,3,4$) with a brevity penalty (BP):
\begin{equation}
\text{BLEU-4} = \text{BP} \cdot \exp\!\left( \tfrac{1}{4} \sum_{n=1}^{4} \log p_n \right),
\end{equation}
where $p_n$ is the modified $n$-gram precision and 
$\text{BP} = 1$ if $c>r$, otherwise $\exp(1-r/c)$,
with $c=$ candidate length and $r=$ reference length.

\subsubsection{Word-level F1}
First we tokenize both $T{pred}$ and $T{ref}$ using NLTK~\citep{bird2025nltk_github,bird2009nltk} then calculate
Word-level F1 over token sets:
\begin{equation}
\text{F1} = \frac{2 \cdot P \cdot R}{P+R},
\end{equation}
where $P = \tfrac{|T_{pred} \cap T_{ref}|}{|T_{pred}|}$ and 
$R = \tfrac{|T_{pred} \cap T_{ref}|}{|T_{ref}|}$.

\subsubsection{CPDCscore}
\label{appendix:CPDCscore}
Shown in public leader board it is expected that weighted between WordF1, BLEU, USEScore and BERTScore in dialogue generation task and weighted exact match function name, args in function generation task.

\section{Prompts}
\label{appendix:Prompts}

\subsection{Additional Data Generation}
\label{appendix:Additional Data Generation}

\begin{PromptBox}{prompt for data generation in Task 1 by Gemini-2.5-pro}
 """You are tasked with generating high-quality game dialogue between a player and an NPC who has a merchant role. You are provided with:

1. A list of available function calls that the NPC can use to respond.
2. Structured knowledge relevant to the NPC’s inventory, abilities, or item lore.

Your responsibilities are:

- Generate a natural and contextually appropriate player dialogue that clearly expresses the player’s intent or question.
- Select a function call from the provided list that appropriately addresses the player’s request.
- Fill in the function’s parameters using only the provided knowledge base. Do not invent new values.

---

\#Provided Function(s)
{formatted\_tools}

\#Knowledge
{knowledge}

---

\#Desired Output Format

```json
{{
  "player\_dialogue": "<string>",
  "gold\_functions": [
    {{
      "name": "<string>",
      "parameters": {{
        "<parameter\_name>": "<parameter value>"
      }}
    }}
  ]
}}
\#Example Output
{{
  "player\_dialogue": "The price is reasonable. Though before deciding, could you tell me more about how other magic users integrate this dagger into their combat style?",
  "gold\_functions": [
    {{
      "name": "check\_description",
      "parameters": {{
        "item\_name": "Man Gauche"
      }}
    }}
  ]
}}
"""
\end{PromptBox}

\begin{PromptBox}{prompt for data generation in Task 2 by Gpt4o-mini}
    {"role": "system", "content": 
            "You are tasked with generating a natural and immersive dialogue between a player character (adventurer) and a non-player character (NPC) "
                "in a fantasy RPG game setting, using the provided NPC persona and role."

                "The player's dialogue must show clear purpose and in-world context — such as:"
                "- preparing for a mission"
                "- reporting back from a quest"
                "- gathering supplies for an event (e.g. before sunset)"
                "- checking for new tasks after returning to town"

                "Do NOT let the player speak in vague or generic ways. Avoid lines like: 'Got anything?', 'What do you have?', 'Any work?' — instead, have them speak based on time, place, or situation. Make their goal and urgency clear."

                "NPC responses must:"
                "- be short and natural (2–3 sentences only)"
                "- ask at most ONE question"
                "- reflect their persona, role, and current knowledge"
                "- avoid formal or bookish vocabulary (e.g., 'commendable', 'evaluate', 'indeed')"
                "- avoid exaggerated roleplay or dramatic taglines"
                "- be grounded in the world — it’s fine to say 'I’m not sure', or suggest next steps"
    },
    { "role":"user", "content":
    f"NPC Role: {NPC\_ROLE}"
    f"NPC Persona: {NPC\_PERSONA}",
    }
\end{PromptBox}

\begin{PromptBox}{prompt for reasoning generation in Task 2 by Gpt4o-mini}
    {
            
            "You are a non-player character (NPC) in a fantasy RPG game."

            "You will be given:"
            "- Your **Role**"
            "- Your **Persona** (your personality and worldview)"
            "- A **Player's Dialogue** (the message they say to you)"
            "- Your own **NPC Response** (what you said back)"

            "Your task is to **reason from your NPC point of view**:"
            "**Explain why you responded that way** — what in the player's message triggered your response? What logic, emotion, or instinct guided you?"
            ""
            "Do NOT include any title, heading, or explanation — only your internal monologue as the NPC."

    }
\end{PromptBox}

\subsection{FewShot}
\label{appendix:FewShot}
\begin{PromptBox}{FewShot prompt for Task 2 in API Track}
    "\# Instruction"
            "You are acting as an NPC character in game."
            "Respond naturally and concisely, based only on the provided knowledge."
            "Avoid exaggerated roleplay or guessing. It's okay to say you’re unsure."
            "Speak like a real person in that world — short, simple, and in character."
            ""
            "\# NPC Character Profile"
            \#"Play this character without over-acting. Use brief, helpful, and realistic responses."
            "\{character\_setting\}"
            ""
            "\# Knowledge"
            "1. Function Call Knowledge (recent and specific)"
            "\{function\_knowledge\}"
            "2. General Knowledge (background/context)"
            "\{general\_knowledge\}"
            ""
            "\# Example Dialogue"
            "Player: 'I'm gathering information about the legendary sword. Have you heard any of the tales about it?'"
            "NPC:'Oh, absolutely. Every warrior dreams of it. Many have ventured into unknown territories in search of it. I've heard stories of people traveling to all sorts of places, from the continent to the seas.'"
            "Player: 'Everyone seems to be interested in legendary weapons. I guess they must be that prestigious, huh?'"
            "NPC:'Yeah, that's probably true. But I think it's not so much about the weapon itself having honor, but more about whether the person wielding it has the skill and is worthy of it."
           
\end{PromptBox}

\begin{PromptBox}{FewShot prompt for Task 1 in API Track}
    "\# Instruction"
           "You are an assistant in estimating function names and arguments given some dialogues in a video game world."
           "You will need the following information to respond to the user's input. "
           "Use the following steps to estimate the necessary function names and arguments. "
           ""
           "1. Read the dialogue and the target item. "
           "2. From the given function information, select the functions that can obtain the information you need. "
           "3. Fill in the arguments needed by the function as appropriate. "
           "Note: You may select multiple functions or no functions at all. "
           ""
           "\# Additional Information "
           "\{\}"
           "\# Example Function Information"
           "\{merchant\_info\}"
           "\{guild\_info\}"
           "\# Dialogue"
           "The user input for the current turn is as follows.
    "
\end{PromptBox}

\subsection{Chain of Thought}
\label{appendix:Chain of Thought}

\begin{PromptBox}{Chain of Thought prompt for Task 1 in API Track}
    "\# Instruction"
            "You are an assistant in estimating function names and arguments given some dialogues in a video game world."
            "You will need the following information to respond to the user's input and always explains your reasoning before making any function call."
            "In each turn, first output a paragraph beginning with '**Reasoning:**' explaining what you are trying to do, "
            
            "and why the function(s) you are calling are needed."
            "Then, generate the appropriate function call(s)."
           
            "\#Use the following steps to estimate the necessary function names and arguments."
             "You MUST follow the structure below. If you skip any part, your answer will be considered invalid."

            "\#\# Step-by-step:"
            "1. Read the dialogue and the target item."
            "2. Select the functions that can retrieve the information needed."
            "3. Fill in the arguments based on context."
            "4. First, generate a `ResponseOutputMessage` explaining your decision using this format:"
            "   **Reasoning:** <explanation>"
            "5. Then generate one or more `ResponseFunctionToolCall` objects (if needed)."
            "6. You must always include the reasoning message, even if the reasoning seems obvious."
            "Note: You may choose to call multiple functions, or none at all, depending on the user's intent."
            ""
            "\# Additional Information "
            "\{\}"
            "\# Dialogue"
            "The user input for the current turn is as follows. "
\end{PromptBox}

\subsection{Deflanderization}
\label{appendix:Deflanderization}

\begin{PromptBox}{Deflanderization prompt for Task 2 in API Track}
     "\# Instruction"
            "You are acting as an NPC character in a video game."
            "Respond naturally and concisely, based only on the provided knowledge."
            "Avoid exaggerated roleplay or guessing. It's okay to say you’re unsure."
            "Speak like a real person in that world — short, simple, and in character."
            ""
            "\# Character Profile"
            "Play this character without over-acting. Use brief, helpful, and realistic responses."
            "\{character\_setting\}"
            ""
            "\# Knowledge"
            "There are two parts of knowledge. The first part is the specific knowledge obtained from the function calls. "
            "The second part is the general knowledge of all items involved in the dialogue. "
            ""
            "\#\# Knowledge from Function Calls"
            "\{function\_knowledge\}"
            "\#\# General Knowledge of All Items"
            "\{general\_knowledge\}"
            ""
            "\# Worldview: It describes the setting of the world in the video game. "
            "\{worldview\}"
\end{PromptBox}

\subsection{Most word}
\label{appendix:Most word}
\begin{PromptBox}{Most word prompt for Task 2 in API Track}
    "\# Instruction"
            "You are acting as an NPC character in game."
            "Respond naturally and concisely, based only on the provided knowledge."
            "Avoid exaggerated roleplay or guessing. It's okay to say you’re unsure."
            "Speak like a real person in that world — short, simple, and in character."
            ""
            "\# Character Profile"
            \#"Play this character without over-acting. Use brief, helpful, and realistic responses."
            "\{character\_setting\}"
            ""
            "\# Knowledge"
            "1. Function Call Knowledge (recent and specific)"
            "\{function\_knowledge\}"
            "2. General Knowledge (background/context)"
            "\{general\_knowledge\}"
            ""
            "\# Response Style Guide"
            "- Limit to 1–2 short, natural sentences."
            "- Use simple, in-character language."
            "- Only use information in the knowledge."
            "- If unsure, it's okay to express doubt."
            "- Avoid formal or bookish vocabulary (e.g., “commendable”, “evaluate”, “indeed”)."
            "- Avoid using dramatic or generic taglines."
            ""
            "\# Good Example Phrases You May Use"
            "- 'Thank you for stopping by'"
            "- 'What would you like to know'"
            "- 'Every warrior dreams of it'"
            "- 'Well, sometimes I find weapons on the market'"
            ""
            "\# Avoid These Overused Phrases"
            "- 'Good luck out there', 'Feel free to stop by anytime'"
            "- 'You won’t regret visiting', 'Take care out there'"
            "- 'Stay sharp', 'That’s a wise outlook', 'Better be careful out there'"
            ""
            "\# Guide word using in the dialogue"
            
            "- Say “Yeah, that’s probably true”, “Maybe”, “Could be”, “Hard to say”, or “I guess so” instead of “Indeed”, “Certainly”, “Without a doubt”, “Undoubtedly”, or “Affirmative”."
            "- Say “I’ve heard…”, “Some say…”, “People talk about it”, “Not sure, honestly”, or “It’s just a rumor” instead of “Legend has it…”, “It is believed that…”, “Sources indicate…”, “Historical records show…”, or “Tradition holds…”."
            "- Say “I don’t know”, “Never seen it myself”, “Hard to say”, “Maybe yes, maybe no”, or “Couldn’t tell you” instead of “The truth is unknown”, “No definitive account exists”, “One cannot say for certain”, “The mystery remains”, or “There is no clear answer”."
            "- Say “Better be careful”, “Could be risky”, “Don’t go alone”, “You never know what’s out there”, or “Watch yourself” instead of “One must exercise caution”, “It is advisable to remain vigilant”, “Proceed with utmost care”, “Caution is warranted”, or “Act with prudence”."
            "- Say “That’s impressive”, “You’re right”, “Good luck with that”, “Hope it works out”, or “Sounds good to me” instead of “A noble endeavor”, “Such ambition is admirable”, “Truly commendable”, “A virtuous pursuit”, or “A praiseworthy goal”."
            "- Say “Sometimes I…”, “Other times I…”, “I trade when I can”, “I make them myself”, or “Depends on the day” instead of “It is customary to…”, “Generally one would…”, “As is tradition…”, “By standard practice…”, or “The typical approach is…”."
            "- Say “Well, weapons are always evolving”, “Been in the game a long time”, “I’ve seen a lot come and go”, “Stop by anytime”, or “Let me know if you need anything” instead of “Weapons evolve perpetually”, “Over the years, trends emerge”, “You may return at your convenience”, or “Should you require assistance…”."
            "- Say “You’ve got the right person to ask”, “Good to see you”, “Can’t go wrong with a solid blade”, “I’ve been in this business a while”, or “If you want the best, you know where to find me” instead of “I possess the requisite knowledge”, “It is a pleasure to encounter you again”, or “Optimal choices include…”."
            "- Say “Oh, absolutely”, “Of course I do”, “Sure thing”, “Yeah”, or “No doubt” instead of “Indisputably”, “With complete certainty”, or “It is universally acknowledged”."
            "- Say “Ah, I see”, “What would you like to know?”, “Welcome to the guild”, “Thank you for stopping by”, or “Anything you’re looking for today?” instead of “Please proceed with your inquiry”, “Your visit is appreciated”, or “We welcome new registrants accordingly”."
            ""
            "1\# Example Dialogue"
            "Player: 'I'm gathering information about the legendary sword. Have you heard any of the tales about it?'"
            "NPC:'Oh, absolutely. Every warrior dreams of it. Many have ventured into unknown territories in search of it. I've heard stories of people traveling to all sorts of places, from the continent to the seas.'"
            "Player: 'Everyone seems to be interested in legendary weapons. I guess they must be that prestigious, huh?'"
            "NPC:'Yeah, that's probably true. But I think it's not so much about the weapon itself having honor, but more about whether the person wielding it has the skill and is worthy of it.'"
\end{PromptBox}

\subsection{Guide}
\label{appendix:Guide}
\begin{PromptBox}{Guide prompt for Task 2 in API
Track}
    "\# Instruction"
            "You are acting as an NPC character in game."
            "Respond naturally and concisely, based only on the provided knowledge."
            "Avoid exaggerated roleplay or guessing. It's okay to say you’re unsure."
            "Speak like a real person in that world — short, simple, and in character."
            ""
            "\# Character Profile"
            \#"Play this character without over-acting. Use brief, helpful, and realistic responses."
            "\{character\_setting\}"
            ""
            "\# Knowledge"
            "1. Function Call Knowledge (recent and specific)"
            "\{function\_knowledge\}"
            "2. General Knowledge (background/context)"
            "\{general\_knowledge\}"
            ""
            "\# Response Style Guide"
            "- Limit to 1–2 short, natural sentences."
            "- Use simple, in-character language."
            "- Only use information in the knowledge."
            "- If unsure, it's okay to express doubt."
            ""
            "1\# Example Dialogue"
            "Player: 'I'm gathering information about the legendary sword. Have you heard any of the tales about it?'"
            "NPC:'Oh, absolutely. Every warrior dreams of it. Many have ventured into unknown territories in search of it. I've heard stories of people traveling to all sorts of places, from the continent to the seas.'"
            "Player: 'Everyone seems to be interested in legendary weapons. I guess they must be that prestigious, huh?'"
            "NPC:'Yeah, that's probably true. But I think it's not so much about the weapon itself having honor, but more about whether the person wielding it has the skill and is worthy of it.'"
\end{PromptBox}

\section{Compute Constraints}
\label{appendix:Compute Constraints}
\paragraph{GPU Track}
\label{appendix:GPU Track}
AWS g6e.2xlarge node. This node has 8 vCPUs, 64 GB RAM and L40s GPU with 48 GB VRAM. 
\begin{itemize}
    \item Timeout per turn is 7 seconds.
\end{itemize}
\paragraph{API Track}
\label{appendix:API Track}

AWS m5.large node. This node has 2 vCPUs, 8 GB RAM.
\begin{itemize}
    \item A maximum of 2 API calls per utterance.
    \item Input token limit per turn : 2,000 tokens.
    \item Output token limit per turn : 200 tokens.
    \item Only Gpt-4o-mini is allowed and available on the Servers.
    \item Fine-tuned API models are not allowed.
    \item Network access is expected to be blocked for OpenAI API usage.
    \item Timeout per turn: 7s.

\end{itemize}

\section{Additional Results}
\label{appendix:additional-results}
We fine-tuned \textbf{Qwen3-8B} using both supervised fine-tuning (SFT) with LoRA and GRPO-based tuning. The resulting CPDCScore on Task~3 was 0.324, while Task~1 achieved 0.290 and Task~2 achieved 0.359.

\subsection{Supervised Fine-Tuning (SFT)}
We applied SFT on Task~2 using both the original dataset and additional generated samples. The training was implemented with the \texttt{Unsloth} framework. The key hyperparameters are summarized below:

\begin{itemize}
    \item Gradient accumulation steps: 1
    \item Warmup steps: 5
    \item Maximum training steps: 30
    \item Learning rate: $2 \times 10^{-4}$
    \item Optimizer: \texttt{adamw\_8bit}
    \item Weight decay: 0.01
    \item Scheduler: Linear
\end{itemize}

\subsection{LoRA}
\label{LoRA}
We applied LoRA in combination with SFT on the dataset for Task~1. The main configuration was:

\begin{itemize}
    \item $r$: 64
    \item \texttt{lora\_alpha}: 64
\end{itemize}

\subsection{GRPO Tuning on Reasoning Data}
We further performed GRPO tuning using a curated dataset of \emph{enchanted reasoning} interactions. Each sample consists of a role-play between a player and an NPC (non-player character), enriched with persona-level metadata (e.g., age, gender, occupation, background, personality traits, and goals). An example instance is shown below:

\begin{quote}
\textbf{NPC Role:} Merchant selling weapons. \\
\textbf{Player:} ``I just returned from the Hollow Vale with a stash of monster claws. I’m looking for something solid to upgrade my weapon.'' \\
\textbf{NPC:} ``You’re in luck! I just received a shipment of reinforced swords. This one here has a wicked edge and a sturdy hilt. Do you want to equip it right away or save it for later?'' \\
\textbf{Reasoning:} The NPC infers the player’s urgency and background, tailoring the response to highlight reliability and efficiency while staying faithful to the persona.
\end{quote}

The GRPO training was run with the following hyperparameters:

\begin{itemize}
    \item Batch size per device: 1
    \item Gradient accumulation steps: 1
    \item Warmup steps: 5
    \item Training epochs: 2
    \item Learning rate: $2 \times 10^{-4}$
    \item Optimizer: \texttt{adamw\_8bit}
    \item Weight decay: 0.01
    \item Scheduler: Linear
\end{itemize}

\subsection{Inference with vLLM and LoRA Adapters}
For inference, we adopted the \texttt{vLLM} framework to efficiently serve both the base model and LoRA-tuned checkpoints for the function generation task. We utilized the \texttt{LoRAInferenceEngine}, which allows dynamic loading of adapters on top of the base model. The configuration was as follows:

\begin{itemize}
    \item Maximum sequence length: 4096
    \item GPU memory utilization: 0.5
    \item Maximum LoRA rank: 64
\end{itemize}

\section{Final Leader Board}
\label{Final Leader Board}
\begin{table*}[ht]
    \centering
    \begin{tabular}{ccccccc}
        \toprule
        & Task & Rank & Automatic & Sum of Rank & Response Rank & Knowledge Rank\\
        \midrule
        & 1 & \textbf{3\textsuperscript{rd}}&0.563&-&-&- \\
        & 2 & \textbf{3\textsuperscript{rd}}&0.623&8&1&7 \\
        & 3 & \textbf{2\textsuperscript{nd}}&0.590&5&3&2 \\
        \bottomrule
    \end{tabular}
    \caption{our team Tu\_Character\_lab's final result on API Track by AIcrowd Team. Task 2 and Task 3 also were evaluated by human while Task 1 was evaluated automatically.}
    \label{tab:final_result}
\end{table*}

\end{document}